\pdfoutput=1
\documentclass[11pt]{article}

\usepackage[]{ACL2023}

\usepackage{times}
\usepackage{latexsym}

\usepackage[T1]{fontenc}
\usepackage[utf8]{inputenc}

\usepackage{microtype}

\usepackage{inconsolata}
\usepackage{diagbox}
\usepackage{graphicx}
\usepackage{verbatim}
\usepackage{multirow}
\usepackage{algorithmic}
\usepackage{longtable}
\usepackage{array}
\usepackage{subfigure}
\usepackage{stfloats}
\usepackage{balance}
\usepackage{multicol}
\usepackage{color}
\usepackage{epstopdf}
\usepackage{bm}
\usepackage{amsmath}
\usepackage{amssymb}
\usepackage{booktabs} % For formal tables
\usepackage{epsfig}
\usepackage{enumitem}
\usepackage{cleveref}
\usepackage{arydshln}
\usepackage{balance}
\usepackage{xspace}
\usepackage{enumitem}

\title{
Building Interpretable and Reliable Open Information Retriever for New Domains Overnight
}

\author{Xiaodong Yu\thanks{\indent Equal Contribution.} \enspace\enspace\enspace Ben Zhou$^{*}$ \enspace\enspace\enspace Dan Roth \\
University of Pennsylvania\\
\tt \{xdyu, xyzhou, danroth\}@seas.upenn.edu}

\begin{document}
\maketitle

\begin{abstract}
% \rewrite{
% Open-domain question answering (QA) can be viewed as a process that first understands the question, finds corresponding world knowledge, and finally composes the knowledge to derive an answer. During the whole process, the step of knowledge retrieval is the most challenging step for many benchmarks, as it requires succinctness, completeness, and correctness. 
Information retrieval (IR) or knowledge retrieval, is a critical component for many downstream tasks such as open-domain question answering (QA). It is also very challenging, as it requires succinctness, completeness, and correctness.
In recent works, dense retrieval models have achieved state-of-the-art (SOTA) performance on in-domain IR and QA benchmarks by representing queries and knowledge passages with dense vectors and learning the lexical and semantic similarity. However, using single dense vectors and end-to-end supervision are not always optimal because queries may require attention to multiple aspects and event implicit knowledge. 
% many queries may need reasoning between several passages, and different target passages may be similar to different parts of the query. 
In this work, we propose an information retrieval pipeline that uses entity/event linking model and query decomposition model to focus more accurately on different information units of the query. We show that, while being more interpretable and reliable, our proposed pipeline significantly improves passage coverages and denotation accuracies across five IR and QA benchmarks. It will be the go-to system to use for applications that need to perform IR on a new domain without much dedicated effort, because of its superior interpretability and cross-domain performance. \footnote{Data and code are available here: \url{https://github.com/CogComp/decomp-el-retriever}.}
% }

% We propose a knowledge retrieval pipeline that improves on all three dimensions through two main novelties: question decomposition and event linking. Specifically, we employ a question decomposition model that parses the question to relevant events, representing the model's educated guess on how the question may be solved if an ideal situation exists. Due to the limitation of language models, such decomposed events are often ``imagined'' and contradict the real world. This is where our second novelty comes in: we use a novel event-linking model that understands the artificial events and finds the real-world events that are as close as possible for a replacement. If the real-world events do not contribute to a solution to the question, our proposed pipeline will continue sampling model guesses until a final answer is found. Compared with existing information retrieval methods such as dense passage retrieval, entity linking, and large pre-trained language models, our solution provides a higher succinctness and completeness due to the question decomposition step and better completeness and correctness because of event linking models' wider coverage at the same time. We show that we improve across several QA benchmarks, namely StrategyQA, HotpotQA, Natural Questions, and TriviaQA. 

\end{abstract}

\section{Introduction}
% \rewrite{
Open-domain information retrieval is a task to retrieve relevant information for any type of queries. An ideal IR model should provide short, accurate and complete information to the downstream models to perform tasks such as open-domain question answering (QA). More importantly, an open-domain IR system should show good cross-dataset performance for it to be truly open-domain, and ideally be interpretable at the same time to provide diverse information and improvement directions for downstream models.
% }
% Open-domain question answering (QA) is a task of understanding queries, retrieving relevant knowledge, and using knowledge to derive the answer. The task is usually designed into two stages: 1) a retriever that aims to understand the query and retrieve relevant passages. 2) a reader that aims to understand the retrieved passages and generate the correct answer. 
Recently, dense retrieval models \cite{karpukhin2020dense, izacard2021contriever} achieves SOTA performances on open-domain QA benchmarks by learning the dense representation of the query and the passages to measure the lexical and semantic similarities. The passages with higher similarity to the query representation will be retrieved. 

\begin{figure}[t]
    \centering
    \includegraphics[scale=0.7]{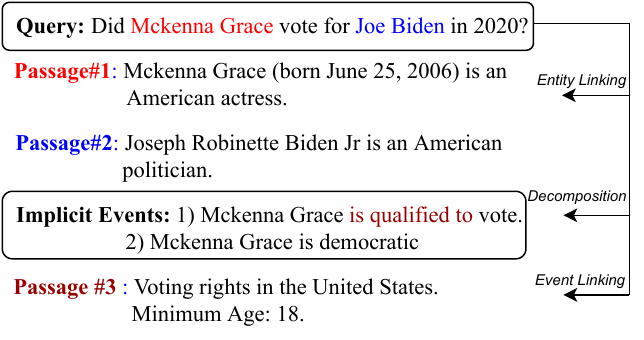}
    \caption{An example of the query that needs reasoning between several passages. }
    \label{fig:example}
\end{figure}

While these approaches achieve relatively high performances when supervised with in-domain training data, they are less ideal to be applied to out-of-domain queries. This is mainly due to their single vector representations and the end-to-end training schemes. For example, in Fig.~\ref{fig:example}, we observe that questions may require symbolic reasoning processes. That is, to answer if McKenna Grace voted for Joe Biden in the 2020 election, we would need to perform information retrieval on implicit dimensions such as the legal age to be allowed to vote and other events that may change Grace's political stances. However, understanding and performing this process is inherently difficult for existing retrievers that rely on single dense vectors \cite{karpukhin2020dense}. Even if they can perform such reasoning processes in a latent way through the representations, it is much less reliable compared with a symbolic process that first explicitly understands what to do and then retrieve relevant information. Furthermore, it is almost impossible to interpret the retrieval process embedded in dense vectors, which hinders research and applications that aim to improve from the mistakes.

% In this paper, we use an entity linking model to retrieve knowledge of entities, and use a query decomposition model to decompose implicit events of the query, and an event linking model to retrieve knowledge of decomposed events. We evaluate our model on 5 different QA benchmarks and show great improvement in passage recall compared with previous dense vector retrievers.

In this paper, we propose an information retrieval pipeline that addresses the aforementioned issues, which tackles questions that require attention to multiple dimensions in an interpretable and reliable fashion. Specifically, our pipeline first tries to decompose the input query into relevant dimensions and associate it with hypothetical events that may contribute to the final answer. With the relevant events, we build on top of the advances of event-linking models \cite{Yu2021EventLG} and link the hypothetical events to real-world information from a traceable knowledge base. Compared with existing methods such as Contriever \cite{izacard2021contriever} and DPR \cite{karpukhin2020dense}, our pipeline does not require any large-scale pre-training, domain-specific training or parameter tuning. It is more interpretable as future work can examine the intermediate generations of hypothetical events and improve individual components based on more fine-grained observations.

At the same time, we show that our pipeline has more reliable cross-domain performances. On a test suite composed of five question-answering datasets, our proposed pipeline improves an average of 6\% compared with the state-of-the-art retriever FiD-KD \cite{Izacard2020DistillingKF} in cross-domain supervision settings, and over 10\% compared with the unsupervised but heavily pre-trained Contriever model. Our proposed pipeline is the best-performing method, while being interpretable and reliable, for any applications that need to perform information retrieval overnight without much in-domain efforts and considerations.

\section{Related Work}
Traditionally, people use sparse vector methods, like TF-IDF and BM25, to calculate the lexical similarity between queries and documents. Though BM25 is still one of the best unsupervised retrieving methods, the lack of semantic understanding limits its performance. After the success of pre-trained language models like BERT \cite{Devlin2019BERTPO}, dense representations show a better semantic understanding of queries and documents. People start to use dense vectors from BERT to train a bi-encoder model as the retriever \cite{karpukhin2020dense} or a cross-encoder model as the ranker \cite{Nogueira2019PassageRW}. Many works use different methods trying to improve the retrieval performance of the bi-encoder. Just to name a few: \newcite{karpukhin2020dense} use annotated Question Answering pairs of queries and documents to train a bi-encoder model with hard negatives mined from BM25. \newcite{Guu2020REALMRL} combine the bi-encoder training into the pre-training of BERT. \newcite{izacard2021contriever} propose an unsupervised bi-encoder retriever by sampling two spans within the same document to create positive pairs. Though different works focus on different parts of the bi-encoder training, using a single vector to represent all the information of the query remains the same, and the effort is always to make the dense vector better understand the query. In our work, we want to point out that encoding the query into a single vector is less optimal, and we aim to use different modules to retrieve passages that are relevant to specific information units of the query, which makes the retrieval process more interpretable and reliable.

\section{Method}
\begin{figure}[t]
    \centering
    \includegraphics[scale=0.35]{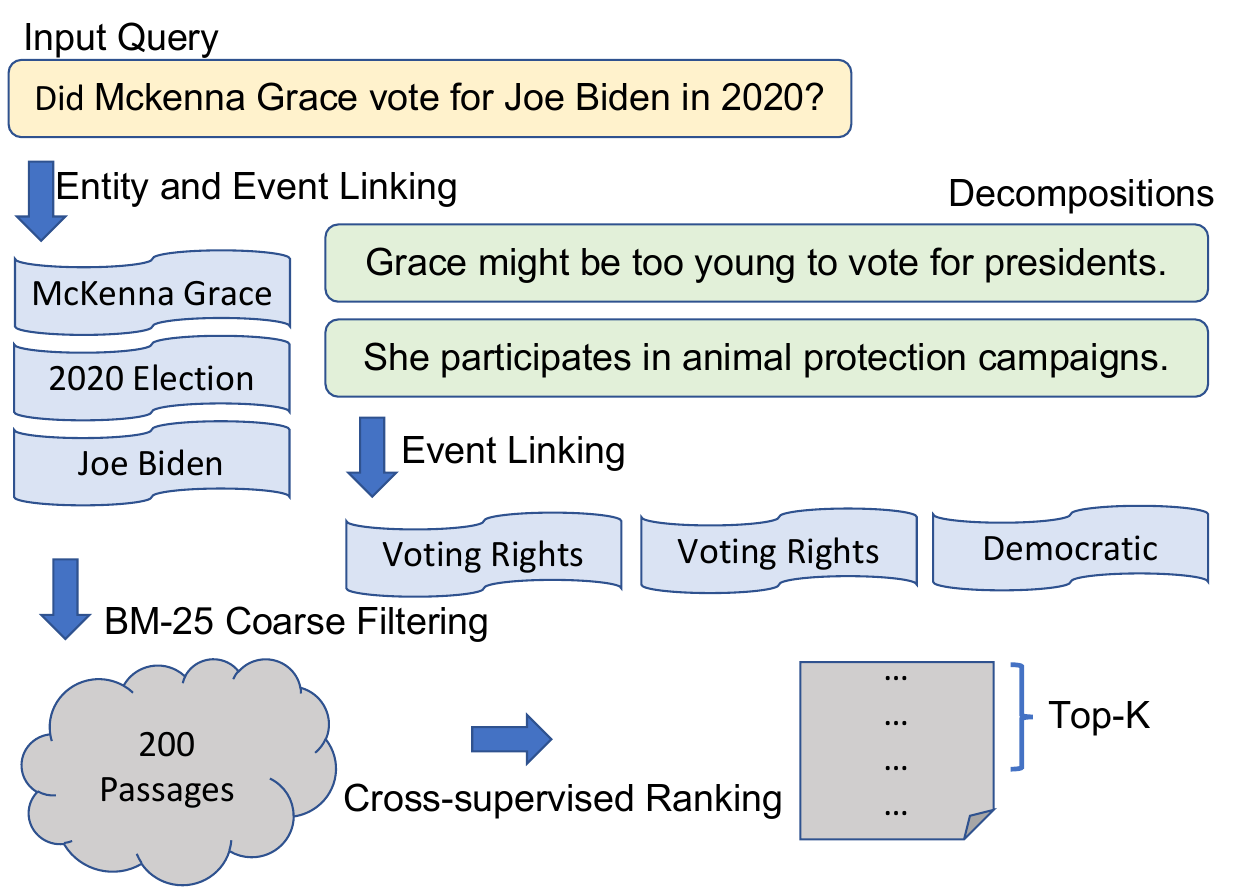}
    \caption{
        An overview of our proposed pipeline system. The input query gets decomposed to relevant events, which are then linked to the knowledge base with an event linking model. The resulting pages go through a fast coarse filtering, followed by a cross-domain supervised fine-grained selection. 
    }
    \label{fig:overview}
\end{figure}

In this section, we describe our information retrieval pipeline, as overviewed in Fig.~\ref{fig:overview}.
In general, our information retrieval is a pipeline that can be described as linking -> decomposition -> linking. Specifically, we first conduct entity and event linking on the input query question to retrieve relevant documents of explicit entities and events in the query. Then, we apply a decomposition model on the query to generate potential supporting events and conduct event linking again. The idea is that our core retriever, the event-linking model, works on identifying similar events from a knowledge base. That being said, the decomposition process is designed to associate the query with similar supporting events, which will be properly utilized by the event-linking system. 

% \subsection{Models}

% \noindent
% \textbf{Entity Linking.} We use GENRE \cite{decao2021autoregressive} to process entity linking requests. This entity linking model will find named entities in the input sentence, and link them to corresponding Wikipedia entries. 

% \noindent
% \textbf{Event Linking.} We use the event linking model EveLink proposed in \citet{Yu2021EventLG}. 

% \noindent
% \textbf{Decomposition.} We use the decomposition system proposed in \citet{ZRYR22}. 

\subsection{Title Generation}

\noindent
As shown in Figure \ref{fig:overview}, given input question/query $Q$, we first use the entity linking system GENRE \cite{decao2021autoregressive} to acquire ten titles\footnote{We get ten since this is the default of the model we use.} and the event linking system EveLink \cite{Yu2021EventLG} to acquire another five titles from Wikipedia. Then, we use the decomposition model and follow previous settings \cite{ZRYR22} to generate five sets of decompositions, each containing three sentences. For each of these sentences from decomposition, we use the event linking model to acquire another five titles.\footnote{Five is a magic number, and we do not tune this parameter on target datasets.} We use the joint set of the resulting titles as the starting point for further passage-level information retrieval. Both entity linking model and event linking model are trained on Wikipedia hyperlinks, without any in-domain supervision.

\subsection{Passage Selection}

Since both the entity linking model and event linking model only retrieve Wikipedia titles, instead of a specific passage, to fairly compare with the previous dense retriever, we build a simple passage selection module to select passages from the Wikipedia pages generated in the previous step. 

\noindent
\textbf{Coarse Filtering.} We split Wikipedia pages into DPR-styled text blocks, each with 100 words, and then use BM-25 implemented by \cite{Ma2021ARS} as an efficient method to select 200 passages as candidates. In practice, we observe that 200 passages are almost the same as document-level coverages.

\noindent
\textbf{Fine-grained Ranking.} To re-rank coarse passage selections and get more fine-grained selections, we train a T5-large model based on existing datasets' evidence annotations. Specifically, the model takes an input sequence that is the concatenation of a question and a piece of context. When the context contributes to the solution-finding process, the output sequence will be a positive token, and if the context is irrelevant to the question, the output sequence will be a negative token. We propose two versions of our pipeline that use different source datasets. The first version uses HotpotQA annotations, with the annotated sentences relevant to the question as positive contexts and other randomly selected sentences as negative contexts. We randomly sample the same amount of negative contexts as the positive ones and result in 430k question-context pairs. The second version uses NaturalQuestions as the supervision source. Specifically, we use the hard negatives provided in \citet{karpukhin2020dense} for the NQ dataset. To clarify the comparisons, we down-sample the NQ supervision to the same size as the HotpotQA model. Both versions are trained with 1 epoch and other default parameters set by the Transformers package \cite{wolf-etal-2020-transformers}. 

\section{Experiments}
% In this section, we describe the experiments to showcase our proposed pipeline.

\begin{table}[t]
\centering
\small
{
\begin{tabular}{lcc}
\toprule
Dataset & \#Dev\\
\cmidrule(lr){1-1}\cmidrule(lr){2-2}
NaturalQuestions & 3,610 \\
TriviaQA & 11,313 \\
HotpotQA & 5,126 \\
BoolQA & 3,270 \\
StrategyQA & 229 \\
\bottomrule
\end{tabular}
}
\caption{Dataset Statistics.} 
\label{tab:statistics}
\end{table}

\begin{table*}[t]
\centering
\small
{
\begin{tabular}{lcccccccc}
\toprule
System & NQ@5 & NQ@20 & TQA@5 & TQA@20 & Hotpot@5 & Hotpot@20 & BoolQ Acc. & STQA Acc.\\
\cmidrule(lr){1-1}\cmidrule(lr){2-2}\cmidrule(lr){3-3}\cmidrule(lr){4-4}\cmidrule(lr){5-5}\cmidrule(lr){6-6}\cmidrule(lr){7-7}\cmidrule(lr){8-8}\cmidrule(lr){9-9}
Contriever & 47.8 & 67.8 & 59.4 & 74.2 & 32.5 & 48.2 & 59.8 & 65.1 \\
DPR (NQ) & (68.3) & (80.1) & 56.9 & 69.0 & 32.1 & 44.9 & 63.3 & 65.9 \\
FiD-KD (cross) & 59.4 & \textbf{74.2} & 66.3 & 77.3 & 36.7 & 50.5 & 68.0 & 63.3 \\
\midrule
Ours (Hotpot) & \textbf{60.5} & 72.9 & 72.0 & 78.7 & (55.6) & (62.9) & 64.5 & \textbf{67.2} \\
Ours (NQ) & (69.0) & (76.3) & \textbf{74.8} & \textbf{79.4} & \textbf{55.1} & \textbf{64.5} & \textbf{68.1} & 63.8 \\
\bottomrule
\end{tabular}
}
\caption{IR performances across different datasets. FiD-KD is trained on TQA for NQ evaluation and trained on NQ for other results. () marks results using in-domain IR supervision. Ours (Hotpot) and Ours (NQ) are our proposed pipelines where the fine-grained filtering model is trained on HotpotQA and NaturalQuestions, respectively. \textbf{Bold} numbers are the best results, excluding in-domain supervised ones.} 
\label{tab:recall}
\end{table*}

\subsection{Datasets and Metrics}
We use several QA datasets that require information retrieval (IR) as our test bench. Specifically, we consider NaturalQuestion (NQ) \cite{kwiatkowski-etal-2019-natural}, TriviaQA (TQA) \cite{joshi-etal-2017-triviaqa}, HotpotQA (Hotpot) \cite{yang-etal-2018-hotpotqa}, BoolQA \cite{clark-etal-2019-boolq} and StrategyQA \cite{geva-etal-2021-aristotle}. Following previous work \cite{izacard2021contriever}, we use the official development sets of each dataset for evaluation purposes. As we do not use any parameters that are specifically tuned or adjusted according to benchmark performances, we do not separately prepare a development set. Table~\ref{tab:statistics} lists the number of instances we use for evaluation from each dataset. 

\noindent
For the first three datasets, we consider the percentage of answers successfully contained by the top-K retrieved texts, where we denote top-K as data@K (also known as recall@K). We remove all yes/no questions from HotpotQA for accurate recall calculation. For BoolQA and StrategyQA, since the answers are binary (yes/no), we employ an entailment model from \citet{ZRYR22} to derive the final answer and evaluate accuracies. 

\subsection{Baselines}
We consider three IR baselines: Contriever \cite{izacard2021contriever}, DPR \cite{karpukhin2020dense} and FiD-KD \cite{Izacard2020DistillingKF}. To fairly compare different methods, we aim to report results that are not directly supervised with in-domain training data. Contriever is an unsupervised model. DPR is trained on NQ. For FiD-KD, since there are multiple released models, we report cross-dataset performances using the TQA-trained system for NQ, and the NQ-trained system for all other datasets. \footnote{We use DPR and FiD-KD implemented by \cite{Lin_etal_SIGIR2021_Pyserini}.}

\begin{table}[t]
\centering
\small
{
\begin{tabular}{lcc}
\toprule
System & Hotpot@5 & Hotpot@20 \\
\cmidrule(lr){1-1}\cmidrule(lr){2-2}\cmidrule(lr){3-3}
Ours (NQ) & 54.6 & 64.0 \\
Ours (NQ) w/ correction & 61.0 & 69.6 \\
\bottomrule
\end{tabular}
}
\caption{GPT-3 ablation study} 
\label{tab:gpt-ablation}
\end{table}

\subsection{Results}
Table~\ref{tab:recall} shows the IR performances of our proposed pipeline compared to state-of-the-art not-directly-supervised IR systems. Our proposed pipeline outperforms the unsupervised Contriever model by an average of 19\% across three datasets with recall@5 results. Compared with the cross-domain DPR model, our pipeline improves a drastic 18\% on TriviaQA, and 23\% on HotpotQA. Even the more competitive FiD-KD model suffers from the cross-dataset transfer, surpassed by our proposed pipeline for 8\% on TriviaQA, and 18\% on HotpotQA. In general, we show that our proposed pipeline gets better open-domain performances without in-domain supervision and is potentially more reliable for real-world applications. 

\subsection{Boosting with Correction}
Many work use large language models (LLM) to perform knowledge extraction for downstream tasks. However, since LLM generations are not guaranteed to be truthful, we still need to perform a traceable information retrieval for reliability-critical tasks, similar to the pipeline we propose in this work. In order to both utilize the memorization capability of LLMs and produce traceable and trustworthy information, we study if our pipeline can be further improved with LLM correction on the generated decompositions. Specifically, we follow \citet{ZRYR22} and ask GPT-3 to fix any factual errors in the decompositions generated by T5, after which we conduct the same linking and retrieval processes. Table~\ref{tab:gpt-ablation} compares the performances of our pipeline after GPT-3 correction on HotpotQA, the most challenging benchmark in our evaluation suite. We observe a 6\% improvement on both recall@5 and recall@20. This shows the versatility of our pipeline, which can be easily improved by the advancements of individual components, instead of proposing new frameworks and spending a considerable amount of resources on training or pre-training each time. 

\section{Conclusion}
In this work, we propose an information retrieval (IR) pipeline built on the advances of query decomposition and event linking. Specifically, we associate the input query with generated events relevant to the query through the decomposition step and then link these events to real-world facts in a knowledge base with an event-linking model. We show that with simple passage selection through BM-25 and cross-domain QA supervision, our pipeline outperforms existing state-of-the-art unsupervised or cross-domain IR models on five datasets by an average of 6\%. Compared with existing methods, our pipeline does not involve heavy pre-training, parameter tuning, or domain-specific supervision. It is an easy-to-use method that achieves reliable and stable performances without much effort for any new domain. It can also be used with large language models to retrieve faithful and traceable information that is even more accurate.

\section{Limitations}
\begin{itemize}
    \item Though both entity linking and event linking models do not use any human-annotated data, our passage selection model is still supervised. How to build an unsupervised but effection passage selection model could be an interesting future work.
    \item In this work, we only focus on the retriever without a reader. Combining our retrieval pipeline with a reader to evaluate the final exact match accuracy could be a good QA system. 
\end{itemize}

\bibliography{anthology,custom}
\bibliographystyle{acl_natbib}

\appendix

\section{Implementations}
\label{sec:implement}
We use 4 Nvidia RTX A6000 48GB GPUs for model training and evaluation.

\end{document}